\documentclass[runningheads]{llncs}
\usepackage[T1]{fontenc}
\usepackage{graphicx,verbatim}
\usepackage{booktabs}
\usepackage[table]{xcolor}
\usepackage{multirow}
\usepackage{array}
\usepackage{makecell}
\usepackage{tabularx}
\usepackage{bm}
\usepackage{makecell}
\usepackage{float}
\usepackage{tcolorbox}
\usepackage{enumitem}
\usepackage{marvosym}

\newcommand*\rot{\rotatebox{55}}

\usepackage[colorlinks,urlcolor=blue,citecolor=blue]{hyperref}

\definecolor{gold}{HTML}{FFD700}
\definecolor{codeblue}{rgb}{0.25, 0.5, 0.5}
\definecolor{codekw}{rgb}{0.35, 0.35, 0.75}
\definecolor{lightblue}{HTML}{DCF2FD}
\definecolor{skyblue}{HTML}{92C5DE}
\definecolor{myblue}{HTML}{6691CD}
\definecolor{mynewblue}{HTML}{dce5f5}
\definecolor{cellmicroscopycolor}{HTML}{FBE2D5} 
\definecolor{breastimagingcolor}{HTML}{DAE9F8}  
\definecolor{chestxraycolor}{HTML}{C0F1C8}     
\definecolor{fundoscopycolor}{HTML}{FAF6A5}     
\definecolor{retinaloctcolor}{HTML}{D9D9D9}    

\definecolor{highlightcolor}{RGB}{215, 221, 225}

\newtcolorbox{HighlighterBox}[2][]{
    arc=3.8pt,
    left=3.0pt,
    right=3.0pt,
    bottom=3pt,
    top=3pt, 
    colback=mynewblue!10,
    colframe=mynewblue!75,
    boxrule=0.8pt,
    colbacktitle=mynewblue!75,
    coltitle=myblue!20!black,
    title=\textbf{#2},
    fonttitle=\bfseries,
    before upper=\justifying,
    #1,
    toptitle=3pt,
    bottomtitle=3pt,
}

\usepackage{ragged2e}

\begin{document}
\title{Lost in Volume: The CT-SpatialVQA Benchmark for Evaluating Semantic-Spatial Understanding of 3D Medical Vision–Language Models}
%

\author{Mashrafi Monon\inst{1} \and
Umaima Rahman\inst{1,2} \and
Asif Hanif\inst{1} \and
Numan Saeed\inst{1}\textsuperscript{\(\dagger\)} \and
Mohammad Yaqub\inst{1}\textsuperscript{\(\dagger\)}}
\authorrunning{M. Monon et al.}
\titlerunning{CT-SpatialVQA}
\institute{Mohamed Bin Zayed University of Artificial Intelligence, Abu Dhabi, UAE \and
New York University Abu Dhabi, Abu Dhabi, UAE}

  
\maketitle              

\begingroup
\renewcommand{\thefootnote}{\ensuremath{\dagger}}
\footnotetext{These authors jointly supervised this work.}
\endgroup


\begin{abstract}
Recent advances in 3D medical vision–language models have enabled joint reasoning over volumetric images and text, showing strong performance in medical visual question-answering (VQA) and report generation. Despite this progress, it remains unclear whether these models learn spatially grounded anatomy from 3D volumes or rely primarily on learned priors and language correlations. This uncertainty stems from the lack of systematic evaluation of semantic–spatial reasoning in volumetric medical VLMs for clinically reliable decision support. To address this gap, we introduce CT-SpatialVQA, a benchmark designed to evaluate semantic–spatial reasoning in 3D CT data. The benchmark comprises 9077 clinically grounded question-answer (QA) pairs derived directly from 1601 radiology reports and CT volumes, which are validated via a robust LLM-assisted pipeline with a 95\% human consensus agreement rate. Our dataset, requires explicit anatomical localization, laterality awareness, structural comparison, and 3D inter-structure relational reasoning. We also introduce standardized evaluation protocol and benchmark eight 3D medical VLMs, finding severe degradation on semantic-spatial reasoning task, averaging 34\% accuracy and often below random, highlighting the need for deeper integration of volumetric evidence for trustworthy clinical use. The code is available at: \url{https://github.com/BioMedIA-MBZUAI/CT-SpatialVQA}.

\keywords{3D Vision-Language Models (VLMs) \and Semantic-Spatial Reasoning  \and Visual Question Answering \and CT \and Large Language Model (LLM)}

\end{abstract}
\section{Introduction}
On a typical day in radiology, a clinician reviews a chest CT to stage a suspected lung cancer. The scan contains hundreds of axial slices, each showing only a small part of the anatomy. As the radiologist scrolls through the volume, a small lesion appears and disappears near the mediastinum. To understand it, the clinician must mentally reconstruct the 3D structure: Is the lesion in front of or behind the great vessels? Does it extend to the pleura, or is it confined within the lung tissue?
Answering these questions requires integrating information across many slices and forming a coherent 3D picture of the anatomy. These spatial judgments are not ancillary; instead, they directly govern diagnosis and procedural planning. Given this fundamental role of spatial reasoning in volumetric imaging, an important question arises: \textit{Can vision-language models (VLMs) develop similar capabilities for understanding 3D spatial representations?} \\

\begin{figure}[t]
    \centering
    \includegraphics[width=\linewidth]{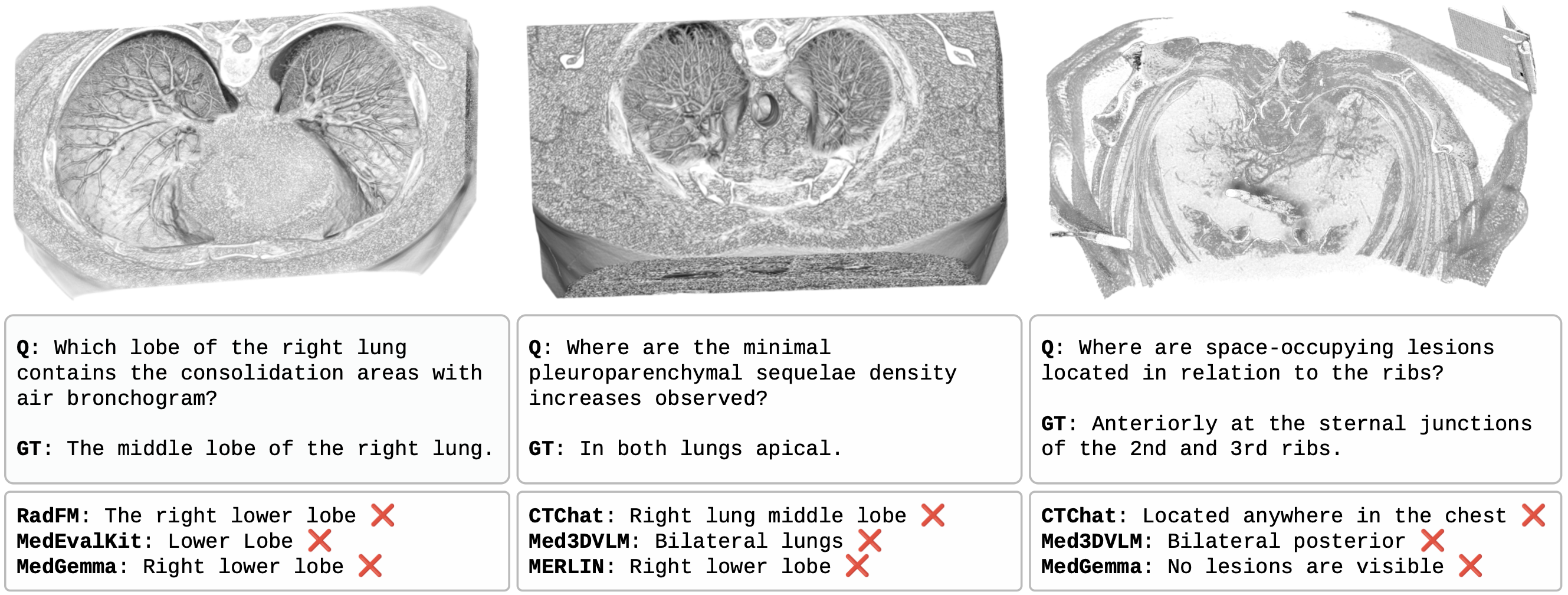}
    \caption{Current 3D medical VLMs struggle to answer spatially grounded questions, highlighting limitations in spatial reasoning and raising concerns about their reliability for clinically meaningful use.}
    \label{fig:ct_spatialvqa_responses}
\end{figure}

\noindent Building on the success of 2D medical vision-language models, recent works have extended it to volumetric medical data and, in some cases, explicitly constructed 3D models designed to encode cross-slice context and spatial structure. This has led to the development of 3D medical VLMs, such as Med3DVLM \cite{med3dvlm}, MERLIN \cite{merlin}, M3D \cite{m3d}, RadFM \cite{radfm}, CT-CLIP, and CT-CHAT \cite{ct-rate}. While these developments raise expectations that 3D medical VLMs may support spatially grounded reasoning, we directly evaluate whether current models fulfill this promise. Through our study, we find that although 3D medical VLMs generate fluent descriptions and correctly answer many semantic or measurement-based questions, their performance deteriorates when tasks demand precise and consistent spatially grounded reasoning, as evident from Figure  \ref{fig:ct_spatialvqa_responses}. This gap underscores a critical distinction: linguistic fluency does not imply true spatial understanding.
Benchmarks such as DeepTumorVQA \cite{deeptumorvqa} primarily emphasize descriptive or measurement-focused queries and provide limited assessment of relational semantic-spatial reasoning. Further analyses of models including M3D \cite{m3d} and RadFM \cite{radfm} demonstrate performance degradation on queries requiring fine-grained 3D structural understanding or stable object–object relationships.\\

\noindent Most VLMs are pretrained on large-scale image text corpora derived from reports, where spatial relationships are weakly specified and often linguistically ambiguous. Recognizing these limitations, recent research in vision–language domain has begun to enhance spatial reasoning.
Within the medical domain \cite{imam2025t3,rahman2025decoupling,rahman2025can,hanif2024baple} however, explicit and systematic investigation of spatial reasoning remains comparatively limited. For instance, recent work in 2D medical imaging \cite{yourotherleft} investigated spatial understanding, showing that models struggle with even simple positional questions such as “Is the left kidney below the stomach?” This issue is only compounded in 3D volumetric imaging, where spatial relationships must be inferred across depth and preserved globally, rather than within a single projection. Even when 3D backbones are used, the lack of explicit mechanisms for enforcing cross-slice consistency limits the models' ability to form a coherent representation of 3D space. While recent efforts like VILA-M3 \cite{vila}, FILA \cite{fila}, E3D-GPT \cite{e3d}, MS-VLM \cite{msvlm}, and BrainMD \cite{brainmd} incorporate expert models or alternative supervision to strengthen anatomical discrimination, they still lack explicit geometric inductive biases and report limitations in precise spatial awareness. Spatial reasoning failures may be amplified in 3D settings, highlighting the need for systematic evaluation in volumetric medical imaging \cite{hanif2023frequency,hanif2025frequency,khan2024guardian}.\\

\noindent Through \textbf{CT-SpatialVQA}, a benchmark designed to evaluate semantic–spatial grounding in 3D medical VLMs, we identify a clear mismatch between current training objectives and the demands of clinical spatial reasoning. Thus, suggesting current training objectives and evaluations predominantly reward semantic fluency, while fine-grained spatial relationships, cross-slice consistency, and volumetric structure remain weakly enforced. \\

\begin{HighlighterBox}{Contributions}
{\footnotesize
\begin{enumerate}[itemsep=3pt]
\vspace{0.2em}
\item \textbf{CT-SpatialVQA.} We introduce CT-SpatialVQA, a benchmark for systematically evaluating semantic-spatial reasoning in 3D medical VLMs using volumetric CT data and clinically grounded QA pairs.
\item \textbf{Spatially-Grounded QA Design.} We generate clinically grounded QA pairs from radiology reports through an LLM-based pipeline comprising generation, validation, and human auditing. The resulting questions require explicit anatomical localization, laterality awareness, structural comparison, and 3D inter-structure reasoning.
\item \textbf{Comprehensive Evaluation.} We evaluate eight 3D medical VLMs on CT-SpatialVQA, quantifying their semantic–spatial reasoning capabilities and identifying failure modes that reveal over-reliance on prior knowledge instead of volumetric visual grounding. We publicly release the CT-SpatialVQA dataset and evaluation code on github.
\end{enumerate}
}
\end{HighlighterBox}

\begin{figure}[t]
    \centering
    \includegraphics[width=\linewidth]{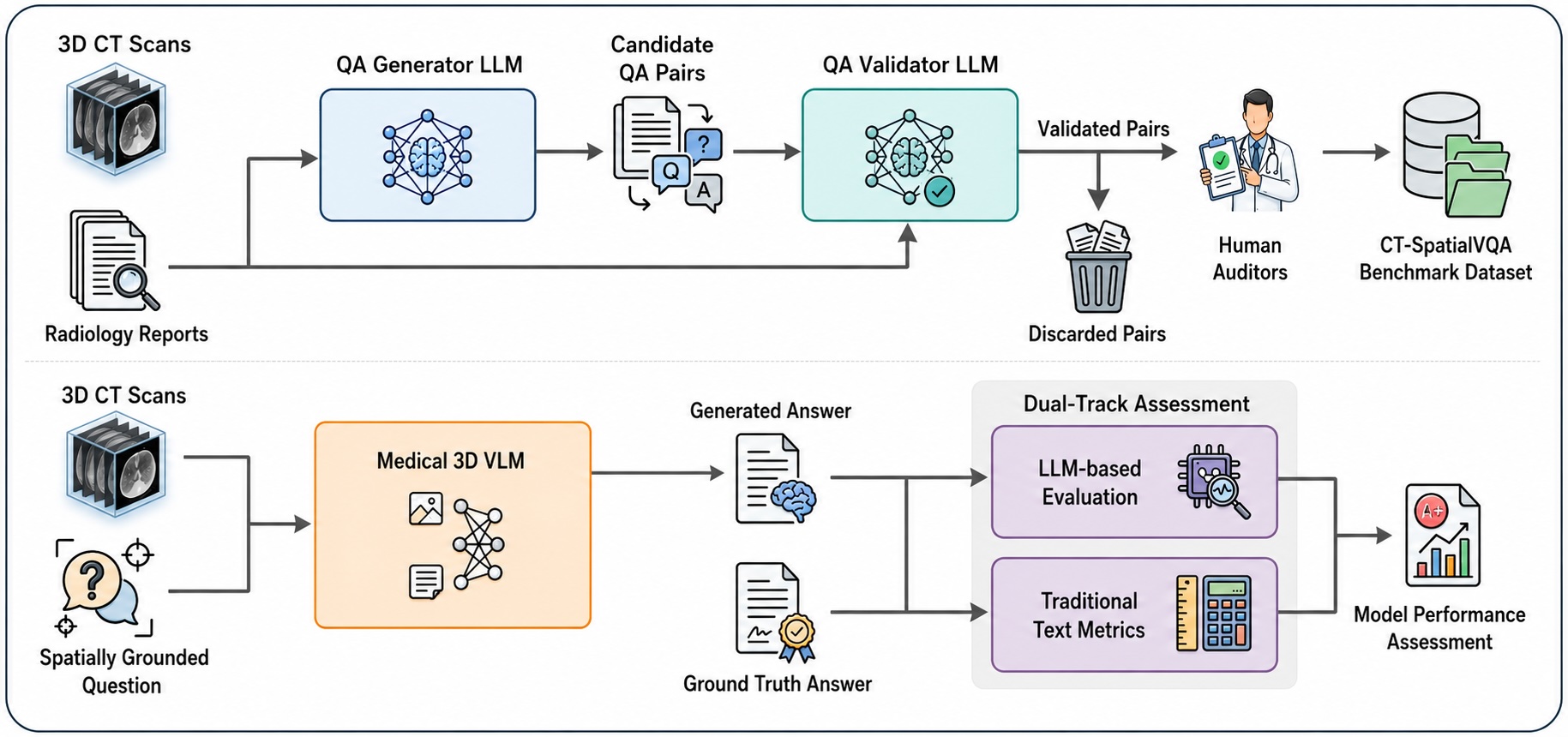}
    \caption{\textbf{An Overview of CT-SpatialVQA} (\textit{top}) The dataset generation pipeline uses paired radiology reports annotated by expert radiologists from 3D CT scans to prompt a "QA Generator LLM", synthesizing candidate spatially-grounded QA pairs. An independent "QA Validator LLM" filters the candidates, and a small subset of remaining pairs is then verified  by human auditors to produce the final benchmark dataset. (\textit{bottom}) The 3D medical VLM evaluation protocol tests a given model by inputting a 3D CT scan and a spatially-grounded question. The model’s generated answer is compared to the ground truth using a dual-track assessment system, where LLM-based evaluation and traditional text metrics are conducted independently.}
    \label{fig:ct_spatialvqa_overview}
\end{figure}

\section{CT-SpatialVQA Benchmark}
\label{sec:ct_spatial_vqa}
The CT-SpatialVQA benchmark is structured in two sequential phases: (1) spatially grounded QA pair generation from radiology reports, and (2) systematic evaluation of 3D medical VLMs using the generated QA pairs to assess semantic–spatial reasoning over volumetric CT data. Figure \ref{fig:ct_spatialvqa_overview} provides an overview of the dataset generation and evaluation phases.\\

\noindent \textbf{Radiology Text Reports.}
The CT-SpatialVQA benchmark is constructed from the CT-RATE \cite{ct-rate} dataset, which consists of volumetric (3D) CT scans paired with radiology reports. These reports are narrative text documents written by board-certified radiologists after reviewing CT volumes. Radiology reports encode rich spatial information describing anatomical structures and pathological findings, including their location, laterality, spatial extent, and relationships to surrounding structures. Spatial descriptions may be explicit (e.g., left lower lobe, bilateral basal atelectasis) or implicit (e.g., apical opacity, central lesion).\\

\begin{figure}[t]
    \centering
    \includegraphics[width=1.0\linewidth]{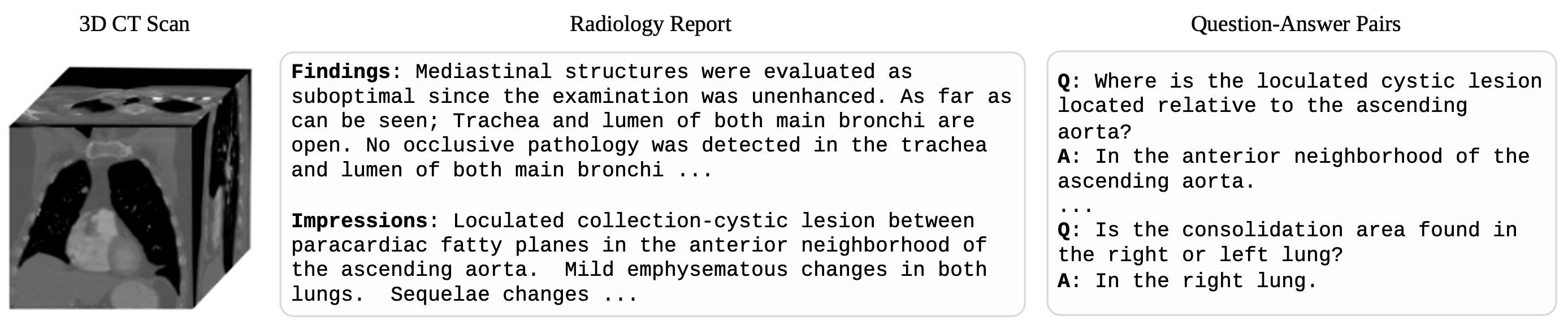}
   \caption{An illustrative example of a 3D CT scan, its corresponding full radiology report, and representative QA pairs.}
    \label{fig:ct_overview}
\end{figure}

\noindent \textbf{Spatially-Grounded Question–Answer Generation.} To construct spatially-grounded QA pairs, each radiology report is provided to an LLM with a carefully designed prompt. The LLM is instructed to identify spatially grounded observations, generate questions that explicitly target semantic-spatial attributes, and provide concise answers derived directly from the report text, while avoiding diagnostic or severity-related content. QA generation focuses exclusively on six spatial dimensions: laterality (left/right/bilateral), longitudinal position (superior/inferior), anterior–posterior relations (front/back orientation), medial–lateral orientation (central/peripheral), adjacency or containment (within/adjacent to structures), and spatial extent or boundaries (confined to or extending across regions). For each identified spatial observation, the LLM generates one or more QA pairs as shown in Figure \ref{fig:ct_overview}, each targeting a single spatial concept with a short, unambiguous answer grounded in the report. The resulting candidate QA pairs are subsequently passed to a validation stage to ensure correctness and reliability.

\begin{HighlighterBox}{\small Instruction Prompt for the QA Generator LLM}
\ttfamily\scriptsize 
\noindent You are a medical AI assistant specialized in radiology and 3D spatial reasoning. Read the 3D CT scan report below and generate 7–10 question–answer (QA) pairs that test a vision–language model’s understanding of spatial and anatomical relationships explicitly described in the report. \\

\noindent Focus only on spatial facts such as:\\
~~~-- Laterality (left/right/bilateral)\\
~~~-- Longitudinal position (superior/inferior)\\
~~~-- Anterior–posterior relations (front/back orientation)\\
~~~-- Medial–lateral orientation (central/peripheral)\\
~~~-- Adjacency or containment (within/adjacent to structures)\\
~~~-- Spatial extent or boundaries (regional confinement or extension)\\

\noindent Guidelines:\\
~~~-- Use only information from the Findings and Impressions sections\\
~~~-- Do not include diagnostic, interpretive, or normality statements\\
~~~-- Questions must emphasize where, which side, above/below, or extent\\
~~~-- Answers must be strictly factual and directly derived from report\\

\noindent \#\#\# Radiology Report\\
\{Findings\}\{Impressions\}
\end{HighlighterBox}

\noindent \textbf{Spatially-Grounded Question–Answer Validation.} To ensure correctness and proper grounding, each candidate QA pair is evaluated together with its corresponding full radiology report by a second, independent LLM acting as a validator. Providing both the generated QA pair and the original report enables the validator to explicitly verify that the answer is directly supported by the report text and that no information is introduced beyond what is stated. The validator further ensures that the question strictly concerns semantic-spatial attributes (e.g., anatomical location, laterality, relative position) and that the answer accurately reflects the spatial relationships described in the report. It also checks for hallucinations, unsupported inferences, ambiguity, and verbosity, ensuring concise and unambiguous responses. QA pairs that fail any of these criteria are discarded.  This two-stage LLM pipeline, combining generation with independent grounding and spatial-consistency validation, reduces hallucinations and improves spatial faithfulness.\\

\begin{HighlighterBox}{\small Instruction Prompt for the QA Validator LLM}
\ttfamily\scriptsize 
\noindent You are a radiology QA validator. You receive radiology report (Findings \& Impressions) and QA pairs created from them. Label each QA pair with two booleans.\\\\
\noindent
~~~~-- is\_spatial: true only if answering the question requires spatial reasoning about anatomical location, orientation, relative position, or laterality visible in the image (e.g., which lung, which lobe, proximity, direction, comparison of organ sizes, location of effusion). Pure presence/absence or textual metadata questions are false.\\
~~~~-- is\_relevant: true only if the question could be answered from the imaging-derived Findings/Impressions (not from general knowledge or the wording of the text itself). Questions that just ask which words appear in the report, cite textual phrases, or hallucinate unseen details are false.\\\\
\noindent \#\#\# Radiology Report\\
\{Findings\}\{Impressions\}\\\\
\noindent \#\#\# QA Pairs\\
\{question\_answer\_pairs\}\\\\
\noindent For each QA pair, return \{'is\_spatial':, 'is\_relevant':\} as JSON only.
\end{HighlighterBox}

\noindent \textbf{Spatially-Grounded Question–Answer Validation (Human Audit).}
To verify the LLM-based validator, we conduct a human audit on a random subset $(12\%)$ of validated QA pairs to assess spatial grounding, correctness, and consistency with the source reports. Trained reviewers with relevant backgrounds examine each QA pair together with the full radiology report for textual support, spatial specificity, correct interpretation, and clinical clarity. We observe 95\% agreement with no systematic discrepancies, indicating that the validator reliably enforces semantic-spatial grounding and aligns closely with human judgment, supporting the robustness of our CT-SpatialVQA benchmark. \\

\noindent \textbf{3D Medical VLM Evaluation.} The finalized QA pairs are used to evaluate 3D medical VLMs. For each sample, a model is provided with a volumetric CT scan and a spatially-grounded question, and it generates an answer based on the 3D image content. Importantly, the corresponding radiology report is not provided during evaluation, ensuring that models cannot rely on textual cues and must instead reason directly from the volumetric data. Performance is assessed through LLM-based evaluation, where external LLMs first act as independent judges and subsequently as a jury to produce binary correctness decisions, alongside traditional text-based similarity metrics.

\begin{figure}[t]
    \centering
    \includegraphics[
        width=\linewidth,
        trim=0 0 0 50, 
        clip
    ]{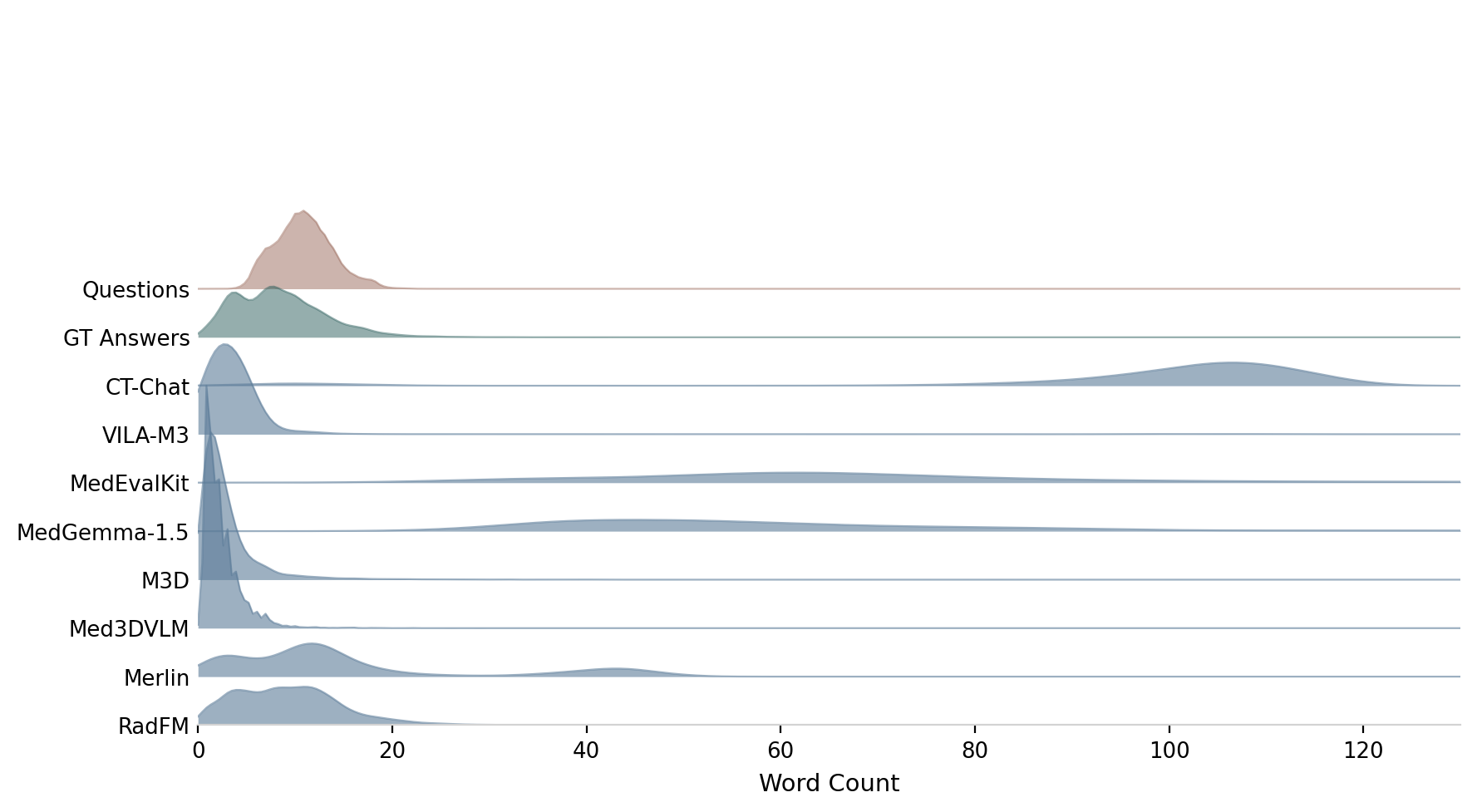}
    \caption{Distribution of word counts in questions, ground-truth answers and predictions of 3D medical VLMs.}
    \label{fig:text_length}
\end{figure}

\section{Experiment and Results}
\label{sec:experiments_results}

\noindent \textbf{3D Medical VLMs.}
To ensure fairness and reproducibility, we restrict our experiments to models with fully accessible implementations and reproducible inference pipelines. We evaluate a diverse set of state-of-the-art 3D medical VLMs, including CT-Chat \cite{ct-rate}, MERLIN \cite{merlin}, Med3DVLM \cite{med3dvlm}, M3D \cite{m3d}, RadFM \cite{radfm}, VILA-M3 \cite{vila}, MedGemma-1.5 (4B) \cite{google2026medgemma}, and MedEvalKit \cite{lingshu}.\\

\noindent \textbf{CT-SpatialVQA Dataset.} We utilized the publicly available CT-RATE \cite{ct-rate} dataset (test split), which contains $1,601$ unique radiology reports paired with their corresponding 3D CT scans. Using \texttt{GPT-4o} as the Generator LLM and \texttt{Gemini 2.5 Flash} as the Validator LLM, we constructed a total of $9,077$ spatially grounded question–answer (QA) pairs. The CT-SpatialVQA dataset contains an average of $5.67$ QA pairs per report. The average length of the questions is $10.89$ words, while the ground-truth answers have an average length of $8.45$ words. Figure \ref{fig:text_length} illustrates the distribution of word counts for questions, ground-truth answers, and predictions generated by eight models. Most models tend to produce concise responses; however, CT-Chat, MedEvalKit and MedGemma generate comparatively longer answers.\\

\noindent \textbf{Inference \& Evaluation Metrics.}
All models are evaluated in a zero-shot setting following their official inference protocols for fairness. 
Volumetric CT inputs are processed using each model’s prescribed preprocessing pipeline (e.g., slice sampling and resizing). We use both reference-based and LLM-based evaluation. GPT-4o \cite{gpt4oapi}, Gemini 2.5 Flash \cite{gemini25flash}, and Qwen3 \cite{qwenplus} act as independent binary judges, with aggregated decisions forming an LLM-as-Jury. Semantic alignment is measured via SBERT cosine similarity \cite{sbert}, and lexical overlap via BLEU \cite{bleu}, ROUGE \cite{rouge}, and METEOR \cite{meteor}.\\

\begin{table}[t]
\centering
\caption{Performance of Medical 3D VLMs on the CT-SpatialVQA Benchmark. Note: The first four rows correspond to Accuracy (\%).}
\label{tab:main_results}
\renewcommand{\arraystretch}{1}
\setlength{\tabcolsep}{3pt}

\resizebox{\textwidth}{!}{
\begin{tabular}{c|llllllll|l}
\toprule
\rowcolor{mynewblue!65}
\cellcolor{mynewblue!65}\shortstack[c]{\textbf{MODELS $\rightarrow$}\\\textbf{METRICS $\downarrow$}}\cellcolor{mynewblue!65}
& \rot{\textbf{CT-CHAT}}
& \rot{\textbf{MERLIN}}
& \rot{\textbf{Med3DVLM}}
& \rot{\textbf{M3D\phantom{ABC}}}
& \rot{\textbf{RadFM\phantom{AB}}}
& \rot{\textbf{VILA-M3}}
& \rot{\textbf{MedGemma}}
& \rot{\textbf{MedEvalKit}}
& \rot{\textbf{AVERAGE}}
\\
\midrule
\midrule
\rowcolor{mynewblue!20} LLM as Judge (Gemini)  & 44.68 & 29.85 & 34.55 & 33.04 & 33.36 & 29.49 & 39.30 & 39.96 & 35.53 \\
\rowcolor{mynewblue!20} LLM as Judge (GPT)  & 41.50 & 29.45 & 31.82 & 31.09 & 31.40 & 28.43 & 40.55 & 39.44 & 34.21 \\
\rowcolor{mynewblue!20} LLM as Judge (Qwen) & 45.27 & 30.34 & 32.20 & 31.21 & 32.09 & 27.81 & 37.33 & 41.89 & 34.79 \\
\rowcolor{mynewblue!20} LLM as Jury         & 43.69 & 28.85 & 31.44 & 30.57 & 31.49 & 28.20 & 38.24 & 40.16 & 34.08\\
\midrule
\rowcolor{mynewblue!20} SBERT Cosine Similarity   & 0.562 & 0.503 & 0.399 & 0.412 & 0.585 & 0.471 & 0.581 & 0.579 & 0.512 \\
\rowcolor{mynewblue!20} BLEU                & 3.290 & 8.110 & 0.560 & 1.080 & 16.45 & 2.510 & 3.070 & 3.460 & 4.816\\
\rowcolor{mynewblue!20} ROUGE-L             & 0.130 & 0.300 & 0.198 & 0.204 & 0.372 & 0.273 & 0.167 & 0.158 & 0.225\\
\rowcolor{mynewblue!20} METEOR              & 0.230 & 0.256 & 0.068 & 0.077 & 0.311 & 0.114 & 0.237 & 0.252 & 0.193\\
\bottomrule
\end{tabular}
}
\end{table}

\begin{figure}[h]
    \centering
    \includegraphics[width=\linewidth]{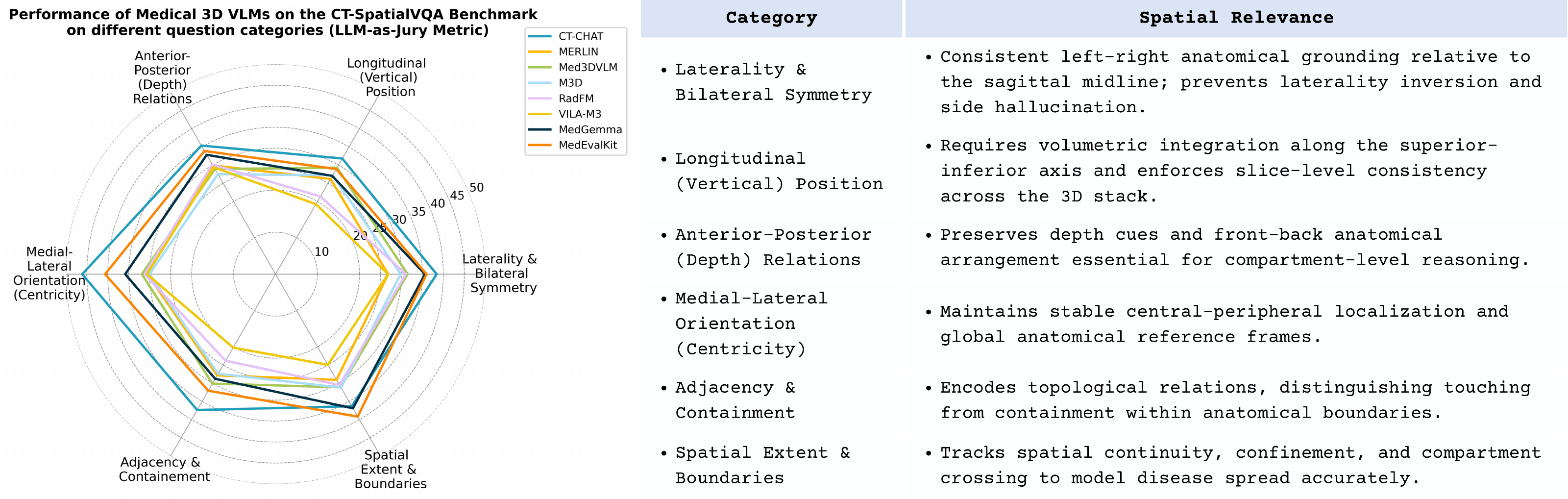}
    \caption{(\textit{left}) Radar plot showing model performance across six spatial reasoning categories in CT-SpatialVQA (LLM-as-Jury). (\textit{right}). Spatial categories considered for our evaluation. Each category captures a clinically relevant spatial primitive required for preserving volumetric spatial integrity in medical VLMs.}
    \label{fig:question_category}
\end{figure}

\noindent \textbf{Results and Discussion.}
Table \ref{tab:main_results} presents a comparison of various 3D medical VLMs on the CT-SpatialVQA dataset. The LLM-as-Jury evaluation indicates that all models struggle to accurately answer spatially grounded questions related to CT scans, with all models falling below 50\% accuracy.  Among the evaluated models, CT-Chat (43.69\%) performs best, followed by MedEvalKit (40.16\%) as the second-best model, and MedGemma-1.5 (38.24\%) in third place. In terms of semantic similarity and lexical overlap metrics, the model demonstrates relatively weak performance on the CT-SpatialVQA dataset. The average SBERT cosine similarity is 0.51, indicating limited semantic alignment between predicted and reference answers. Lexical-based metrics show similarly modest results, with an average BLEU score of 4.81, ROUGE-L (F1) of 0.22, and METEOR of 0.19. Overall, these scores suggest that the generated answers frequently diverge from the ground-truth responses both semantically and lexically, reflecting poor performance on spatial reasoning questions within the dataset.\\

\noindent Figure \ref{fig:question_category} further analyzes performance at the category level. Accuracy varies across spatial relation types, with sharper declines in vertically and depth-oriented reasoning compared to central–peripheral localization. This pattern suggests that performance varies across spatial dimensions, with lower accuracy on relations that require consistent volumetric representations.

\section{Conclusion}
In this work, we introduce \textbf{CT-SpatialVQA}, a clinically grounded benchmark designed to systematically evaluate semantic–spatial reasoning in volumetric CT imaging. CT-SpatialVQA reframes spatial grounding as an explicit and measurable requirement for 3D medical AI. By decomposing spatial reasoning into clinically meaningful primitives and exposing consistent model failures, we establish spatial grounding as a benchmarkable property rather than an assumed byproduct of multimodal scaling. While spatial reasoning limitations are known to affect vision–language models more broadly, our findings demonstrate how these weaknesses manifest in high-stakes 3D clinical settings, where geometric and topological precision are essential. The results highlight structural constraints in prevailing training paradigms that rely heavily on large-scale image–text supervision with limited spatial specificity. CT-SpatialVQA provides a diagnostic framework and research direction, highlighting the need for 3D medical VLMs to move beyond surface correlations toward architectures that encode volumetric structure and spatial consistency.



%
%
%
\bibliographystyle{splncs04}
\bibliography{refs}

@article{ct-rate,
  title={Generalist foundation models from a multimodal dataset for 3D computed tomography},
  author={Hamamci, Ibrahim Ethem and Er, Sezgin and Wang, Chenyu and Almas, Furkan and Simsek, Ayse Gulnihan and Esirgun, Sevval Nil and Dogan, Irem and Durugol, Omer Faruk and Hou, Benjamin and Shit, Suprosanna and others},
  journal={Nature Biomedical Engineering},
  pages={1--19},
  year={2026},
  publisher={Nature Publishing Group UK London}
}

@article{med3dvlm,
  title={Med3dvlm: An efficient vision-language model for 3d medical image analysis},
  author={Xin, Yu and Ates, Gorkem Can and Gong, Kuang and Shao, Wei},
  journal={IEEE Journal of Biomedical and Health Informatics},
  year={2025},
  publisher={IEEE}
}

@article{m3d,
  title={M3d: Advancing 3d medical image analysis with multi-modal large language models},
  author={Bai, Fan and Du, Yuxin and Huang, Tiejun and Meng, Max Q-H and Zhao, Bo},
  journal={arXiv preprint arXiv:2404.00578},
  year={2024}
}

@article{merlin,
  title={Merlin: A vision language foundation model for 3d computed tomography},
  author={Blankemeier, Louis and Cohen, Joseph Paul and Kumar, Ashwin and Van Veen, Dave and Gardezi, Syed Jamal Safdar and Paschali, Magdalini and Chen, Zhihong and Delbrouck, Jean-Benoit and Reis, Eduardo and Truyts, Cesar and others},
  journal={Research Square},
  pages={rs--3},
  year={2024}
}

@article{radfm,
  title={Towards generalist foundation model for radiology by leveraging web-scale 2d\&3d medical data},
  author={Wu, Chaoyi and Zhang, Xiaoman and Zhang, Ya and Hui, Hui and Wang, Yanfeng and Xie, Weidi},
  journal={Nature Communications},
  volume={16},
  number={1},
  pages={7866},
  year={2025},
  publisher={Nature Publishing Group UK London}
}

@inproceedings{vila,
  title={Vila-m3: Enhancing vision-language models with medical expert knowledge},
  author={Nath, Vishwesh and Li, Wenqi and Yang, Dong and Myronenko, Andriy and Zheng, Mingxin and Lu, Yao and Liu, Zhijian and Yin, Hongxu and Law, Yee Man and Tang, Yucheng and others},
  booktitle={Proceedings of the Computer Vision and Pattern Recognition Conference},
  pages={14788--14798},
  year={2025}
}

@article{lingshu,
  title={Lingshu: A generalist foundation model for unified multimodal medical understanding and reasoning},
  author={Xu, Weiwen and Chan, Hou Pong and Li, Long and Aljunied, Mahani and Yuan, Ruifeng and Wang, Jianyu and Xiao, Chenghao and Chen, Guizhen and Liu, Chaoqun and Li, Zhaodonghui and others},
  journal={arXiv preprint arXiv:2506.07044},
  year={2025}
}

@misc{google2026medgemma,
  title        = {MedGemma 1.5 Model Card},
  author       = {{Google Research}},
  year         = {2026},
  url          = {https://huggingface.co/google/medgemma-1.5-4b-it},
  note         = {Accessed: 2026-02-22},
  organization = {Google},
}

@article{deeptumorvqa,
  title={Are vision language models ready for clinical diagnosis? a 3d medical benchmark for tumor-centric visual question answering},
  author={Chen, Yixiong and Xiao, Wenjie and Bassi, Pedro RAS and Zhou, Xinze and Er, Sezgin and Hamamci, Ibrahim Ethem and Zhou, Zongwei and Yuille, Alan},
  journal={arXiv preprint arXiv:2505.18915},
  year={2025}
}

@article{e3d,
  title={E3D-GPT: enhanced 3D visual foundation for medical vision-language model},
  author={Lai, Haoran and Jiang, Zihang and Yao, Qingsong and Wang, Rongsheng and He, Zhiyang and Tao, Xiaodong and Wei, Wei and Lv, Weifu and Zhou, S Kevin},
  journal={arXiv preprint arXiv:2410.14200},
  year={2024}
}

@article{msvlm,
  title={Read like a radiologist: efficient vision-language model for 3D medical imaging interpretation},
  author={Lee, Changsun and Park, Sangjoon and Shin, Cheong-Il and Choi, Woo Hee and Park, Hyun Jeong and Lee, Jeong Eun and Ye, Jong Chul},
  journal={arXiv preprint arXiv:2412.13558},
  year={2024}
}

@article{fila,
  title={Large-scale and fine-grained vision-language pre-training for enhanced ct image understanding},
  author={Shui, Zhongyi and Zhang, Jianpeng and Cao, Weiwei and Wang, Sinuo and Guo, Ruizhe and Lu, Le and Yang, Lin and Ye, Xianghua and Liang, Tingbo and Zhang, Qi and others},
  journal={arXiv preprint arXiv:2501.14548},
  year={2025}
}

@article{brainmd,
  title={Enhancing vision-language models for medical imaging: bridging the 3D gap with innovative slice selection},
  author={Wang, Yuli and Dai, Yuwei and Jones, Craig and Sair, Haris and Shen, Jinglai and Loizou, Nicolas and Hsu, Wen-Chi and Imami, Maliha and Jiao, Zhicheng and Zhang, Paul and others},
  journal={Advances in Neural Information Processing Systems},
  volume={37},
  pages={99947--99964},
  year={2024}
}

@inproceedings{bleu,
  title={Bleu: a method for automatic evaluation of machine translation},
  author={Papineni, Kishore and Roukos, Salim and Ward, Todd and Zhu, Wei-Jing},
  booktitle={Proceedings of the 40th annual meeting of the Association for Computational Linguistics},
  pages={311--318},
  year={2002}
}

@inproceedings{sbert,
  title={Sentence-bert: Sentence embeddings using siamese bert-networks},
  author={Reimers, Nils and Gurevych, Iryna},
  booktitle={Proceedings of the 2019 conference on empirical methods in natural language processing and the 9th international joint conference on natural language processing (EMNLP-IJCNLP)},
  pages={3982--3992},
  year={2019}
}

@inproceedings{rouge,
  title={Rouge: A package for automatic evaluation of summaries},
  author={Lin, Chin-Yew},
  booktitle={Text summarization branches out},
  pages={74--81},
  year={2004}
}

@inproceedings{meteor,
  title={METEOR: An automatic metric for MT evaluation with improved correlation with human judgments},
  author={Banerjee, Satanjeev and Lavie, Alon},
  booktitle={Proceedings of the acl workshop on intrinsic and extrinsic evaluation measures for machine translation and/or summarization},
  pages={65--72},
  year={2005}
}

@misc{gemini25flash,
  title        = {Gemini 2.5 Flash (model code: gemini-2.5-flash)},
  author       = {{Google AI for Developers}},
  howpublished = {\url{https://ai.google.dev/gemini-api/docs/models/gemini-2.5-flash}},
  note         = {Accessed 2026-02-24}
}

@misc{qwenplus,
  title        = {Qwen-Plus (Qwen3 series) model listing},
  author       = {{Alibaba Cloud Model Studio}},
  howpublished = {\url{https://www.alibabacloud.com/help/en/model-studio/models}},
  note         = {Accessed 2026-02-24}
}

@misc{gpt4oapi,
  title        = {GPT-4o model documentation},
  author       = {{OpenAI}},
  howpublished = {\url{https://platform.openai.com/docs/models/gpt-4o}},
  note         = {Accessed 2026-02-24}
}

@misc{yourotherleft,
      title={Your other Left! Vision-Language Models Fail to Identify Relative Positions in Medical Images}, 
      author={Daniel Wolf and Heiko Hillenhagen and Billurvan Taskin and Alex Bäuerle and Meinrad Beer and Michael Götz and Timo Ropinski},
      year={2025},
      eprint={2508.00549},
      archivePrefix={arXiv},
      primaryClass={cs.CV},
      url={https://arxiv.org/abs/2508.00549}, 
}

@article{imam2025t3,
 title={T3: Test-Time Model Merging in VLMs for Zero-Shot Medical Imaging Analysis},
 author={Imam, Raza and Wang, Hu and Mahapatra, Dwarikanath and Yaqub, Mohammad},
 journal={arXiv preprint arXiv:2510.27265},
 year={2025}
}

@inproceedings{rahman2025decoupling,
 title={Decoupling clinical and class-agnostic features for reliable few-shot adaptation under shift},
 author={Rahman, Umaima and Imam, Raza and Yaqub, Mohammad and Mahapatra, Dwarikanath},
 booktitle={International Workshop on Uncertainty for Safe Utilization of Machine Learning in Medical Imaging},
 pages={123--133},
 year={2025},
 organization={Springer}
}

@inproceedings{rahman2025can,
 title={Can language-guided unsupervised adaptation improve medical image classification using unpaired images and texts?},
 author={Rahman, Umaima and Imam, Raza and Yaqub, Mohammad and Amor, Boulbaba Ben and Mahapatra, Dwarikanath},
 booktitle={2025 IEEE 22nd International Symposium on Biomedical Imaging (ISBI)},
 pages={1--5},
 year={2025},
 organization={IEEE}
}

@inproceedings{hanif2024baple,
  title={Baple: Backdoor attacks on medical foundational models using prompt learning},
  author={Hanif, Asif and Shamshad, Fahad and Awais, Muhammad and Naseer, Muzammal and Khan, Fahad Shahbaz and Nandakumar, Karthik and Khan, Salman and Anwer, Rao Muhammad},
  booktitle={International Conference on Medical Image Computing and Computer-Assisted Intervention},
  pages={443--453},
  year={2024},
  organization={Springer}
}

@inproceedings{hanif2025frequency,
  title={On frequency domain adversarial vulnerabilities of volumetric medical image segmentation},
  author={Hanif, Asif and Naseer, Muzammal and Khan, Salman and Khan, Fahad Shahbaz},
  booktitle={2025 IEEE 22nd International Symposium on Biomedical Imaging (ISBI)},
  pages={01--05},
  year={2025},
  organization={IEEE}
}

@inproceedings{hanif2023frequency,
  title={Frequency domain adversarial training for robust volumetric medical segmentation},
  author={Hanif, Asif and Naseer, Muzammal and Khan, Salman and Shah, Mubarak and Khan, Fahad Shahbaz},
  booktitle={International Conference on Medical Image Computing and Computer-Assisted Intervention},
  pages={457--467},
  year={2023},
  organization={Springer}
}

@inproceedings{khan2024guardian,
  title={Guardian: Guarding against uncertainty and adversarial risks in robot-assisted surgeries},
  author={Khan, Ufaq and Nawaz, Umair and Sheikh, Tooba T and Hanif, Asif and Yaqub, Mohammad},
  booktitle={International Workshop on Uncertainty for Safe Utilization of Machine Learning in Medical Imaging},
  pages={59--69},
  year={2024},
  organization={Springer}
}

\end{document}